\documentclass[letterpaper, 10 pt, conference]{sty/ieeeconf}

\usepackage{amsmath,amsfonts,amssymb}
\usepackage{algorithmic}
\usepackage{algorithm}
\usepackage{array}
\usepackage[caption=false,font=normalsize,labelfont=sf,textfont=sf]{subfig}
\usepackage{textcomp}
\usepackage{stfloats}
\usepackage{url}
\usepackage{verbatim}
\usepackage{booktabs}
\usepackage{multirow}
\usepackage{rotating}
\usepackage{makecell}
\usepackage{cite}
\usepackage{xspace}
\usepackage{siunitx}
\usepackage{sty/my_acronyms}
\usepackage{tikz}
\usetikzlibrary{shapes,arrows,calc}
\newcommand{\RomanNumeralCaps}[1] {\MakeUppercase{\romannumeral #1}}

\DeclareSIUnit\milligauss{mG}

\IEEEoverridecommandlockouts

\newcommand{\MAGYC}{\mbox{MAGYC}\xspace}

\newcommand{\MAGYCBFG}{\mbox{MAGYC-BFG}\xspace}
\newcommand{\MAGYCIFG}{\mbox{MAGYC-IFG}\xspace}
\newcommand{\TWOSTEP}{\mbox{TWOSTEP}\xspace}
\newcommand{\MagFactor}{\mbox{MagFactor3}\xspace}

\title{\LARGE \bf
Towards a Factor Graph-Based Method using Angular Rates for Full Magnetometer Calibration and Gyroscope Bias Estimation
}

\author{Sebastián Rodríguez-Martínez$^{1}$ and Giancarlo Troni$^{1}$%
\thanks{This work was supported by the David and Lucile Packard Foundation and FONDECYT-Chile under grant 11180907.}
\thanks{S. Rodríguez-Martínez and G. Troni are with the Monterey Bay Aquarium Research Institute, Moss Landing, CA 95039--9644.
        \{{\tt\small srodriguez, gtroni}\}{\tt\small @mbari.org}}%
}

\begin{document}

\maketitle
\thispagestyle{empty}
\pagestyle{empty}

\begin{abstract}

\acs{mems} Attitude Heading Reference Systems are widely employed to determine a system's attitude, but sensor measurement biases limit their accuracy. This paper introduces a novel factor graph-based method called MAgnetometer and GYroscope Calibration (\MAGYC). \MAGYC leverages three-axis angular rate measurements from an angular rate gyroscope to enhance calibration for batch and online applications. Our approach imposes less restrictive conditions for instrument movements required for calibration, eliminates the need for knowledge of the local magnetic field or instrument attitude, and facilitates integration into factor graph algorithms within Smoothing and Mapping frameworks. We evaluate the proposed methods through numerical simulations and in-field experimental assessments using a sensor installed on an underwater vehicle. Ultimately, our proposed methods reduced the underwater vehicle's heading error standard deviation from \SI{6.21}{\degree} to \SI{0.57}{\degree} for a standard seafloor mapping survey.

\end{abstract}

\acresetall

\section{Introduction}
\label{sec:intro}

%
\ac{mems} \acp{ahrs} are widely used to determine system attitude, primarily in vehicle navigation systems operating in space, ground, and marine scenarios. They play a crucial role in GPS-denied applications, such as autonomous underwater vehicle deployments or navigation in enclosed spaces like tunnels, where inertial navigation is the primary means to estimate system state. Despite their importance, these sensors are susceptible to calibration errors and may not always provide accurate navigation. While expensive high-grade sensors ($>\!100k$ USD) can partially mitigate this issue, addressing calibration errors remains a challenge for mid and low-range sensors designed for widespread use, typically costing thousands of dollars.

A standard \ac{mems} \ac{ahrs} comprises a three-axis magnetometer, a three-axis accelerometer, a three-axis gyroscope, and a temperature sensor. The magnetometer measures the local Earth's magnetic field, aiding in determining the system's heading. The accelerometer, when external accelerations are minimal, measures the system's inclination relative to local gravity, providing pitch and roll orientation information. Lastly, the gyroscope offers the vehicle's angular rate, aiding in refining the rotation estimation.

Accurate estimation hinges on effectively mitigating biases, scale factors, and non-orthogonality issues that can affect \ac{ahrs} components. Gyroscopes and accelerometers face challenges from biases, while magnetometers are also susceptible to calibration errors induced by nearby ferrous materials or electric currents, which can skew the magnetic field, leading to inaccuracies in heading estimation. There are two primary types of magnetometer calibration errors: hard-iron biases, caused by permanent magnetic fields from the vehicle and onboard instruments, resulting in constant output bias, and soft-iron biases, arising from magnetic materials near the sensor, which distort the magnetic field with scale factors and non-orthogonalities.

%
\begin{figure}[t!]
\centering
\begin{tikzpicture}[scale=0.73]
\node[anchor=center,inner sep=0, rotate=33, anchor=center] at (-2.1,-0.03) {\includegraphics[width=3.5cm,height=1.3cm]{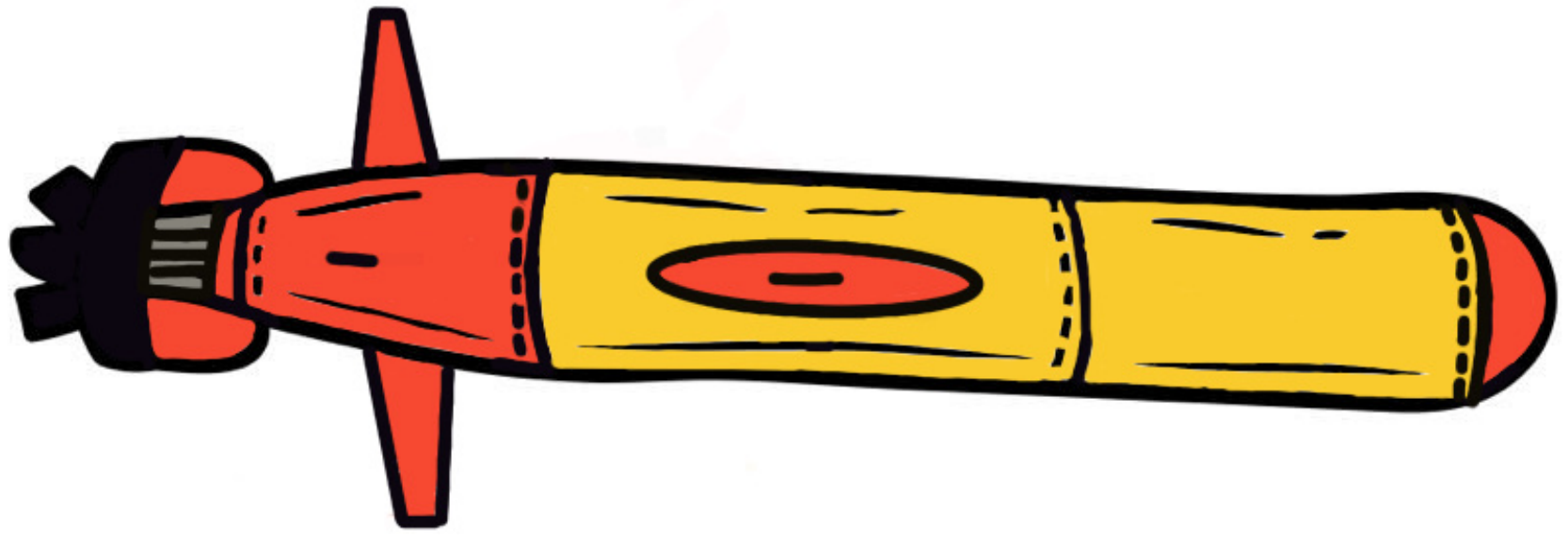}};
\draw [dashed, gray, ultra thick, rotate around={32:(-2,0)}] (-2,0) -- (5.4,0);
\node[rotate=32, gray] at (2.8,3.3) {Ground Truth};
\draw [dashed, black, ultra thick, rotate around={30:(-2,0)}] (-2,0) -- (5.4,0);
\node[rotate=30] at (3.2,2.7) {\MAGYC};
\draw [dashed, red, ultra thick, rotate around={20:(-2,0)}] (-2,0) -- (5,0);
\node[rotate=20, red] at (4.3,2.0) {Raw};
\draw [-, black, ultra thick] (-2,1.2) arc (90:32:1.2);
\draw [->, red, ultra thick] (-2,0) -- (-2,1.8) node[above] {\(\vec{m}\)};
\draw [->, black, ultra thick, rotate around={32:(-2,0)}] (-2,0) -- (1.2,0) node[below] {\(x\) };
\draw [->, black, ultra thick, rotate around={32:(-2,0)}] (-2,0) -- (-2,-1) node[below] {\(y\)};
\node[font=\small] at (-1.2, 1.3) {$\psi$};
\node at (0, -1.6) {Top View};
\end{tikzpicture}
\caption{Diagram illustrating an underwater vehicle's dead reckoning position using different magnetic field sources for the heading estimation with the corresponding trajectories represented with dashed lines.}
\label{fig:attitude_diagram}
\end{figure}
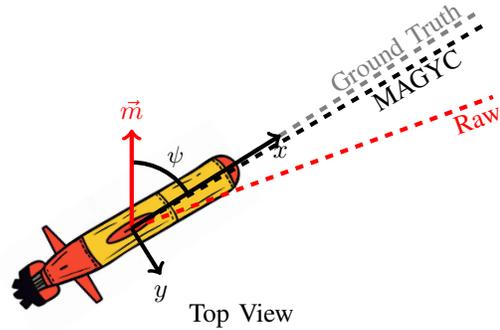

%
The accurate calibration of three-axis magnetometers is essential for reliable attitude estimation. Various methods have been proposed to estimate calibration parameters without the need for additional reference sensors. One widely adopted approach Alonso and Shuster introduced is the \TWOSTEP method. It initially estimates the magnetometer's bias \cite{Alonso2002a}, followed by estimating the scale and non-orthogonality factors as well using an iterative least squares minimization \cite{Alonso2002b}. Several least squares methods have been reported as well \cite{Fang2011, Foster2008, Ousaloo2017}. Another formulation for magnetometer calibration treats it as an ellipsoid fitting problem, which Vasconcelos et al. \cite{Vasconcelos2011} addressed using a maximum likelihood estimate method.

Given the availability of inertial sensors with magnetometers as a package, some strategies combine accelerometer and magnetometer measurements to estimate the magnetometer sensor biases, leveraging accelerometers to measure the local gravity vector \cite{Kok2016, Papafotis2019, Ammann2015}. However, these methods are susceptible to calibration errors caused by translational accelerations. Troni and Whitcomb \cite{Troni2020} introduced a novel approach that uses angular velocity measurements to estimate the magnetometer's hard-iron. Spielvogel and Whitcomb later extended this method to include the estimation of gyroscope and accelerometer biases \cite{Spielvogel2018}. Nevertheless, neither of these methods addresses soft-iron magnetometer calibration.

The methods mentioned above are limited to batch calibrations, where all the measurements must be collected in advance. However, it is desirable to perform calibration in the field when pre-calibration is not possible, for instance, due to changes in the vehicle's configuration deployed in highly disturbed environments. Crassidis et al. \cite{crassidis2005} presented an extension to the \TWOSTEP method that incorporates an \ac{ekf} and an \ac{ukf}. Later, Ma and Jing \cite{Ma2005}, as well as Soken and Sakai \cite{Soken2019} proposed alternative approaches for the \ac{ukf} method, while Guo et al. \cite{Guo2008} presented an alternative \ac{ekf} method. Additionally, Han et al. \cite{Han2017} and Spielvogel et al. \cite{Spielvogel2022} proposed a gyroscope-aided \ac{ekf}, with the latter also incorporating gyroscope biases in the estimation.

%
The previously reported approaches have at least one of the following limitations that narrow their implementation scenarios: ($i$) the requirement to undergo extensive angular motion in all three rotational degrees of freedom, which is often infeasible for devices mounted on full-scale vehicles, such as pitch and roll-stable \acp{rov}; ($ii$) the need for accurate knowledge of the local Earth's magnetic field from magnetic field models which may introduce significant errors due to unmodeled local perturbations \cite{NOAA2023}, or ($iii$) the inability to determine the magnetometer's soft-iron error, i.e., neglect non-orthogonalities and scale factors that affect systems when ferromagnetic materials, such as iron and steel, surround the object, resulting in less accurate output. This paper introduces a novel approach, called MAgnetometer and GYroscope Calibration (\MAGYC), which effectively overcomes these challenges. The approach enables two factor graph-based methods and offers several notable advantages, including ($i$) fewer restrictions on angular movement requirements, ($ii$) no need for information about the local magnetic field or system's attitude, ($iii$) a complete calibration for both magnetometer and gyroscope, and ($iv$) integration into factor graph algorithms for Smoothing and Mapping frameworks with tools such as GTSAM or iSAM \cite{BORGLab2023} for problems such as \ac{slam}.

The paper is structured as follows. Section \RomanNumeralCaps{2} briefly overviews the mathematical content. Section \RomanNumeralCaps{3} details the proposed \MAGYC methods. Section \RomanNumeralCaps{4} describes the evaluation methodology, while Section \RomanNumeralCaps{5} reports simulation results, and Section \RomanNumeralCaps{6} presents experimental results. The results and conclusions are summarized in Section \RomanNumeralCaps{7}.

\section{Mathematical Background}
\label{sec:math}

\subsection{Rotation Matrix}

The orientation of an instrument frame $V$ relative to an inertial world frame $W$ can be represented by a rotation matrix $\mathbf{R}(t) \in SO(3)$ \cite[Def. 3.1]{Lynch2017}.

%
\subsection{Kronecker Product}

The Kronecker product of the matrix $\mathbf{A} \in \mathbb{M}^{p \times q}$ with the matrix $\mathbf{B} \in \mathbb{M}^{r \times s}$ is defined as an $p \times q$ block matrix whose $(i, j)$ block is the $r \times s$ matrix $a_{ij}\mathbf{B}$ \cite[Def. 2.1]{Schacke2013}.

%
\subsection{Operators}

\subsubsection{Skew-Symmetric}

We denote $\left[ \; \cdot \; \right]_\times: \mathbb{R}^3 \rightarrow \mathbb{R}^{3\times3}$ as the usual skew-symmetric operator \cite[Def. 3.7]{Lynch2017}.

\subsubsection{vec-operator}

We denote $vec(\cdot)$ as the vec-operator for any matrix $\mathbf{A} \in \mathbb{M}^{m \times n}$, representing the entries of $\mathbf{A}$ stacked columnwise, forming a vector of length $m \times n$ \cite[Def. 2.2]{Schacke2013}.

%
\subsection{Factor Graphs}

Factor graphs are representations of probabilistic or graphical models consisting of nodes representing unknown random variables ($x_i \in \mathcal {X} $) and edges representing the dependencies or relationships between these variables. The edges are associated with factors ($f_i \in \mathcal{F}$), which are probabilistic constraints derived from measurements, prior knowledge, or relationships between variables, and can be categorized as unary factors when connecting to a single node or binary factors if they connect two or more nodes \cite{Dellaert2012, Koller2009, Kschischang2001}.

The factor graph is a concise and intuitive way of representing a model that is well-suited for performing various types of probabilistic inference tasks, such as bias estimation given a set of measurements for sensor calibration. In comparison to traditional filtering methods, factor graphs have shown superior capability in managing nonlinear processes and measurement models. They offer notable advantages in processing time, flexibility, and modularity, particularly for large-scale and complex optimization problems \cite{Dellaert2006, Dai2022}.

Unary and binary factors can be represented as measurement likelihood with a Gaussian noise model \cite{Dellaert2006}, which is only evaluated as a function of the unknown state $\mathbf{x}$ since the measurement $m$ is considered known
\begin{equation}
L(\mathbf{x}; m) = \exp \left\{ -\frac{1}{2} ||h(\mathbf{x}) - m||_\mathbf{\Sigma}^2 \right\},
\label{eq: unary_factor}
\end{equation}

\noindent where $h(\mathbf{x})$ is a nonlinear measurement model, $|| \, \cdot \, ||$ is the Mahalanobis distance, and $\mathbf{\Sigma}$ is the noise covariance matrix.

Utilizing the factors and a prior belief over the unknown variable $P(\mathbf{x_0})$, we formulate the joint probability model. Our objective is to determine the optimal set of parameters $\mathbf{X}^*$ through the maximum a posteriori (MAP) estimate by maximizing the joint probability $P(\mathbf{X}, \mathbf{Z})$, leading to a nonlinear least squares problem (\ref{eq: factor_map}). Leveraging the factor graph structure and sparse connections in this estimation process helps reduce computational complexity \cite{Dellaert2006, Dellaert2012}.
\begin{equation}
\mathbf{X}^* \triangleq \arg\!\max_{x} \,P (\mathcal{X} | \mathcal{Z}) =  \arg\!\min_{x} \, -\!\log P(\mathcal{X}, \mathcal{Z})
\label{eq: factor_map}
\end{equation}

\section{Proposed Calibration Approach}
\label{sec:approach}

In this Section, we propose two methods based on the novel \MAGYC method to estimate the complete calibration of a three-axis magnetometer, i.e., hard-iron and soft-iron, and a three-axis gyroscope using magnetometer and angular rate measurements in the instrument frame, i.e., the attitude of the instrument is not required.

%
\subsection{Sensor Error Model}
\label{sec:approach.sen_model}

As detailed in Section \RomanNumeralCaps{1}, magnetometers can exhibit biases referred to as hard-iron and soft-iron during operational conditions, leading to potential inaccuracies in their measurements. These biases are assumed to remain relatively constant or change slowly over time, allowing them to be treated as constants. The magnetometer model is given by
\begin{equation}
\mathbf{m_m}(t) = \mathbf{A}(\mathbf{m_t}(t) + \mathbf{m_b}),
 \label{eq: magnetometer_model}
 \end{equation}

\noindent where $\mathbf{m_m}(t)  \in \mathbb{R}^3$ is the noise-free magnetic field measurement in the sensor's frame, $\mathbf{m_t}(t) \in \mathbb{R}^3$ is the noise-free magnetic field real value in the sensor's frame, $\mathbf{A} \in \mathbb{R}^{3 \times 3}$ is the soft-iron, represented by a constant fully populated positive definite symmetric (PDS) matrix, and $\mathbf{m_b} \in \mathbb{R}^3$ is a constant pseudo-hard-iron, that once scaled by $A$ will give us the magnetometer's hard-iron.

In contrast, gyroscopes are affected by the constant sensor bias and can be represented as
\begin{equation}
\mathbf{w_m}(t) = \mathbf{w_t}(t) + \mathbf{w_b},
 \label{eq: gyroscope_model}
 \end{equation}

\noindent where $\mathbf{w_m}(t)  \in \mathbb{R}^3$ is the noise-free gyroscope measurement in the sensor's frame, $\mathbf{w_t}(t) \in \mathbb{R}^3$ is the noise-free gyroscope real value in the sensor's frame, and $\mathbf{w_b} \in \mathbb{R}^3$ is the constant gyroscope bias.

%
\subsection{System Model}
\label{sec:approach.sys_model}

A magnetometer measures, in instrument coordinates, the Earth’s local magnetic field, which is considered to be \textit{locally} constant and fixed with respect to the inertial world frame of reference. From (\ref{eq: magnetometer_model}), we can clear the true magnetic field ($\mathbf{m_t}(t)$) and convert it to world coordinates given a rotation matrix $\mathbf{R}$. Then, we can differentiate the equation with respect to time, removing the local magnetic field from the system's model, which yields to
\begin{equation}
\dot{\mathbf{R}}(\mathbf{A}^{-1}\mathbf{m_m}(t) - \mathbf{m_b}) + \mathbf{R}\mathbf{A}^{-1}\mathbf{\dot{m}_m}(t) = 0.
\end{equation}

Using the standard equation $\dot{\mathbf{R}}(t) = \mathbf{R}(t)\left[\mathbf{w}(t)\right]_\times$ \cite{Lynch2017}, and incorporating the gyroscope bias, we derive a more comprehensive AHRS calibration that considers both magnetometer and gyroscope calibrations. This yields a nonlinear system model independent of the instrument's attitude $\mathbf{R}(t)$.
\begin{equation}
\left[\mathbf{w_m}(t) - \mathbf{w_b}\right]_\times (\mathbf{A}^{-1}\mathbf{m_m}(t) - \mathbf{m_b}) + \mathbf{A}^{-1}\mathbf{\dot{m}_m}(t) = 0
\label{eq: non_linear_model_gyro_bias}
\end{equation}

From (\ref{eq: non_linear_model_gyro_bias}), we estimate the soft-iron ($\mathbf{A}$), the pseudo-hard-iron ($\mathbf{m_b}$), therefore, the hard-iron ($\mathbf{A}\mathbf{m_b}$), and the gyroscope bias ($\mathbf{w_b}$), using gyroscope ($\mathbf{w_m}(t)$) and magnetometer ($\mathbf{m_m}(t)$) measurements. The proposed approach presents two methods based on a pose graph-based approach. While similar implementations using non-linear least squares were analyzed, the factor graph approach demonstrated superior performance. However, due to space constraints, further details regarding this comparison are not included.

%
\subsection{\MAGYC: MAgnetometer and GYroscope Calibration}
\label{sec:approach.fg}

The proposed method for estimating the calibration parameters adopts a factor graph approach, where a single node represents the state $\mbox{$\mathbf{x} =  [\mathbf{a} \; \mathbf{m_b} \; \mathbf{w_b}]^T$}$, where $\mbox{$\mathbf{a} = ( \mathbf{A}_{00} \; \mathbf{A}_{01} \; \mathbf{A}_{02} \; \mathbf{A}_{11} \; \mathbf{A}_{12} \; \mathbf{A}_{22} )^T$}$ is the vector of the unique six upper triangular terms of the soft-iron matrix, $\mathbf{A}$. The residual from the non-linear model (\ref{eq: non_linear_model_gyro_bias}) and any additional constraints required are represented as unary factors.

The decision to employ a single node instead of multiple, one for each magnetometer-gyroscope measurement set (i.e., $m$ nodes), is based on the assumption of constant calibration parameters. If multiple nodes were used, a binary factor would be necessary between each state, totaling $m - 1$ constraints to maintain the constancy of the state vector; however, this is a soft constraint due to the probabilistic nature of the state vector in a factor graph. Modeling the system as a single node ensures one set of calibration values and spares the use of the $m - 1$ constraints.

As per (\ref{eq: non_linear_model_gyro_bias}), a unary factor can be defined for the residual of each magnetometer-gyroscope measurement pair. The equation for this factor is shown in (\ref{eq: unary_factor_residual}), where $h_{r}(x)$ represents the residual model (\ref{eq: non_linear_model_gyro_bias}).
\begin{equation}
L_R(\mathbf{x};(\mathbf{z}_{mag}, \mathbf{z}_{gyro})) = \exp \left\{ -\frac{1}{2} ||h_{r}(\mathbf{x})||_\mathbf{\Sigma}^2 \right\}
\label{eq: unary_factor_residual}
\end{equation}

To use the unary factor in (\ref{eq: unary_factor_residual}), we need to compute the algebraic Jacobian of the residual model. This can be achieved using the Kronecker product and vec-operator, as shown in (\ref{eq: jacobian_residual_factor}). For conciseness, we do not expand these equations further in this paper. In (\ref{eq: jacobian_residual_factor}), the matrix $\mathbf{C}$ represents the inverse of the soft-iron matrix, the vector $\mathbf{c}$ contains the upper triangular terms of matrix $\mathbf{C}$, and $i$ denotes the $i$th sample. Note that the magnetic field differentiation is not directly available and must be computed numerically.
\begin{subequations}\label{eq: jacobian_residual_factor}
\begin{align}
\frac{h_{ri}(\mathbf{x})}{\mathbf{c}} &=  (\mathbf{m_i}^T \otimes [\mathbf{w_i} - \mathbf{w_b}] + \mathbf{\dot{m}_i}^T \otimes \mathbb{I}_ 3) \frac{\partial vec(\mathbf{C})}{\partial \mathbf{c}} \\
\frac{h_{ri}(\mathbf{x})}{\mathbf{m_b}} &=  [\mathbf{w_b} - \mathbf{w_i}] \\
\frac{h_{ri}(\mathbf{x})}{\mathbf{w_b}} &= -\frac{\partial \left(\mathbf{m_i} \otimes [\mathbf{w_b}]\right)}{\partial \mathbf{w_b}} \cdot vec(\mathbf{C}) +  \frac{\partial [\mathbf{w_b}]}{\partial \mathbf{w_b}} \cdot \mathbf{m_b} 
\end{align}
\end{subequations}

To prevent the algorithm from converging to the trivial solution $\mathbf{x} = \mathbf{0}$, which satisfies (\ref{eq: non_linear_model_gyro_bias}), a constraint can be applied to the unique upper triangular terms of the soft-iron matrix. This constraint is based on prior knowledge of the soft-iron matrix's proximity to the identity matrix. Specifically, we calculate an error term by measuring the difference between the Frobenius norm of the unique soft-iron's upper triangular terms and 1. This error is expressed as $N(\mathbf{x}) = ||\mathbf{c}||_F - 1$. We introduce a soft-iron norm unary factor ($L_N$) defined as
\begin{equation}
L_N(\mathbf{x}) = \exp \left\{ -\frac{1}{2} \left| \left| N(\mathbf{x}) \right| \right|_\mathbf{\Sigma}^2 \right\}.
\label{eq: unary_factor_norm}
\end{equation}

The Jacobian of this factor can be found using
\begin{equation}
\frac{\partial N(\mathbf{x})}{\partial \mathbf{c}} = \frac{\mathbf{c}^T}{|| \, \mathbf{c} \, ||_F}.
\label{eq: jacobian_norm_factor}
\end{equation}

It is worth noting that the same factors introduced in this paper, namely (\ref{eq: unary_factor_residual}) and (\ref{eq: unary_factor_norm}), can be added to different value nodes in the graph to enable biases to vary over time and be integrated with SLAM graphs. However, this integration is beyond the scope of this paper.

In the factor graph construction, given a set of $m$ pairs of magnetometer-gyroscope measurements, the factor graph should be composed of a single node for the state vector and $2m$ factors: $m$ factors from the system model residual \eqref{eq: unary_factor_residual} and $m$ factors enforcing unitary norm constraints on the unique upper triangular terms of the soft-iron matrix \eqref{eq: unary_factor_norm}. Nevertheless, to reduce the computational load of the graph, we employ an averaging window of size $\theta$, which reduces the unary factors from $2m$ to $2n$, with $n = m/\theta$. By averaging $\theta$ measurements before incorporating them into the graph, we can decrease the number of factors and effectively smooth the raw measurements and the movement while filtering out high-frequency noise in the signal, preserving the underlying trend. In this paper, to ensure real-time operation during long-duration tasks, the averaging window's length is set to the sensor's frequency, allowing the factor graph to manage the same period, irrespective of the sensor's frequency.

The factor graph's construction can follow two methods based on implementation scenarios. In the first method, calibration is performed as a post-processing step, where all $m$ measurements are initially collected, and then the optimization process incorporates all $2n$ factors. The second approach involves constructing the graph incrementally, with factors added as they become available, and optimization occurs after a specified number of factors have been added.

As mentioned earlier in Section \RomanNumeralCaps{2}.D, constructing the graph requires solving a nonlinear least squares problem with sparsity properties. To address this challenge, Rosen et al. \cite{Rosen2014} proposed RISE, an incremental trust-region method designed for robust online sparse least-squares estimation. As demonstrated in \cite{Rosen2014}, compared to current state-of-the-art sequential sparse least-squares solvers, RISE offers improved robustness against nonlinearity in the objective function and numerical ill-conditioning, leveraging recent advancements in incremental optimization for fast online computation.
\begin{figure}[!t]
\centering
\subfloat[]{\includegraphics[width=1.05in]{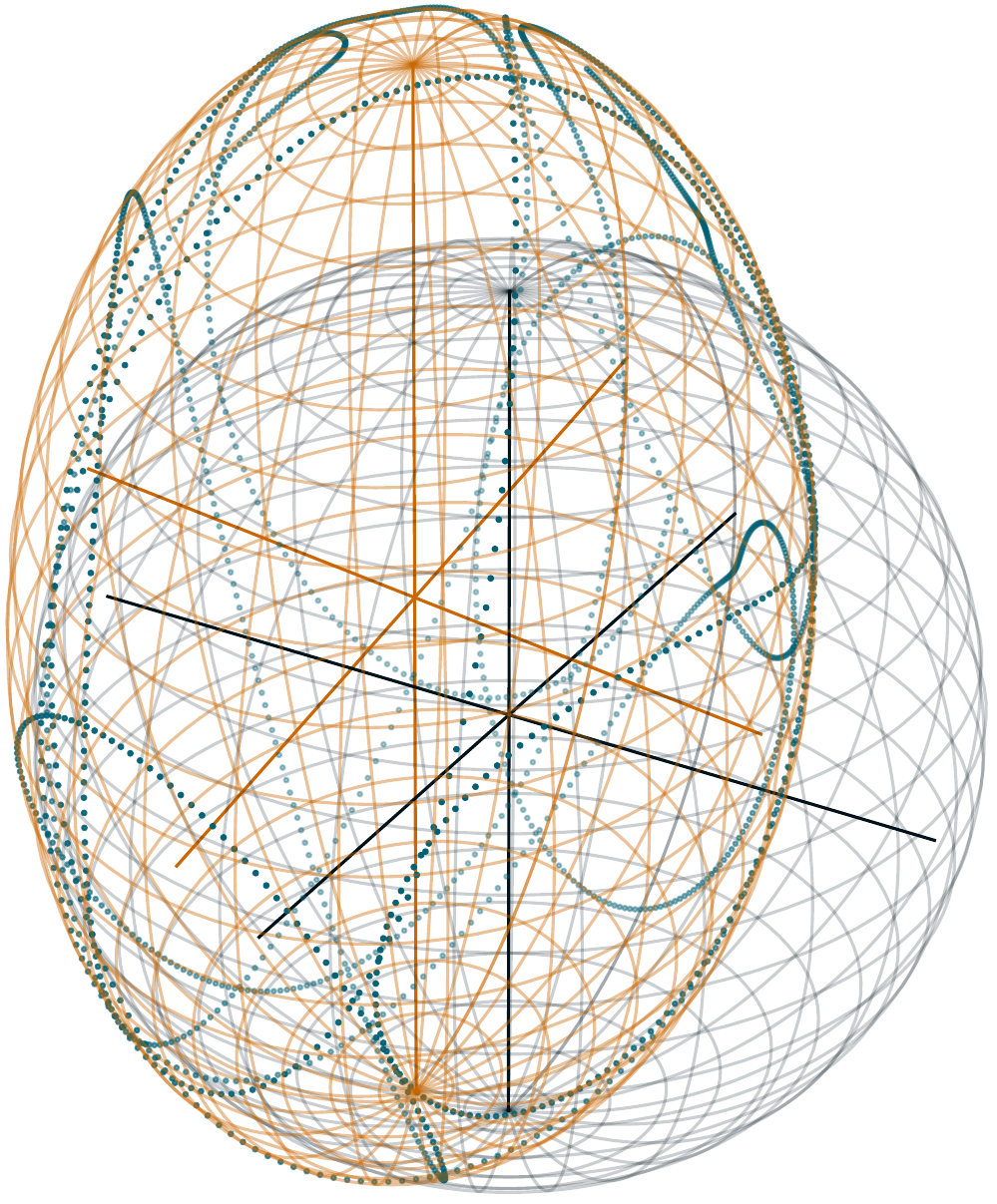}%
\label{fig: sim_data_high_plot}}
\hfil
\subfloat[]{\includegraphics[width=1.05in]{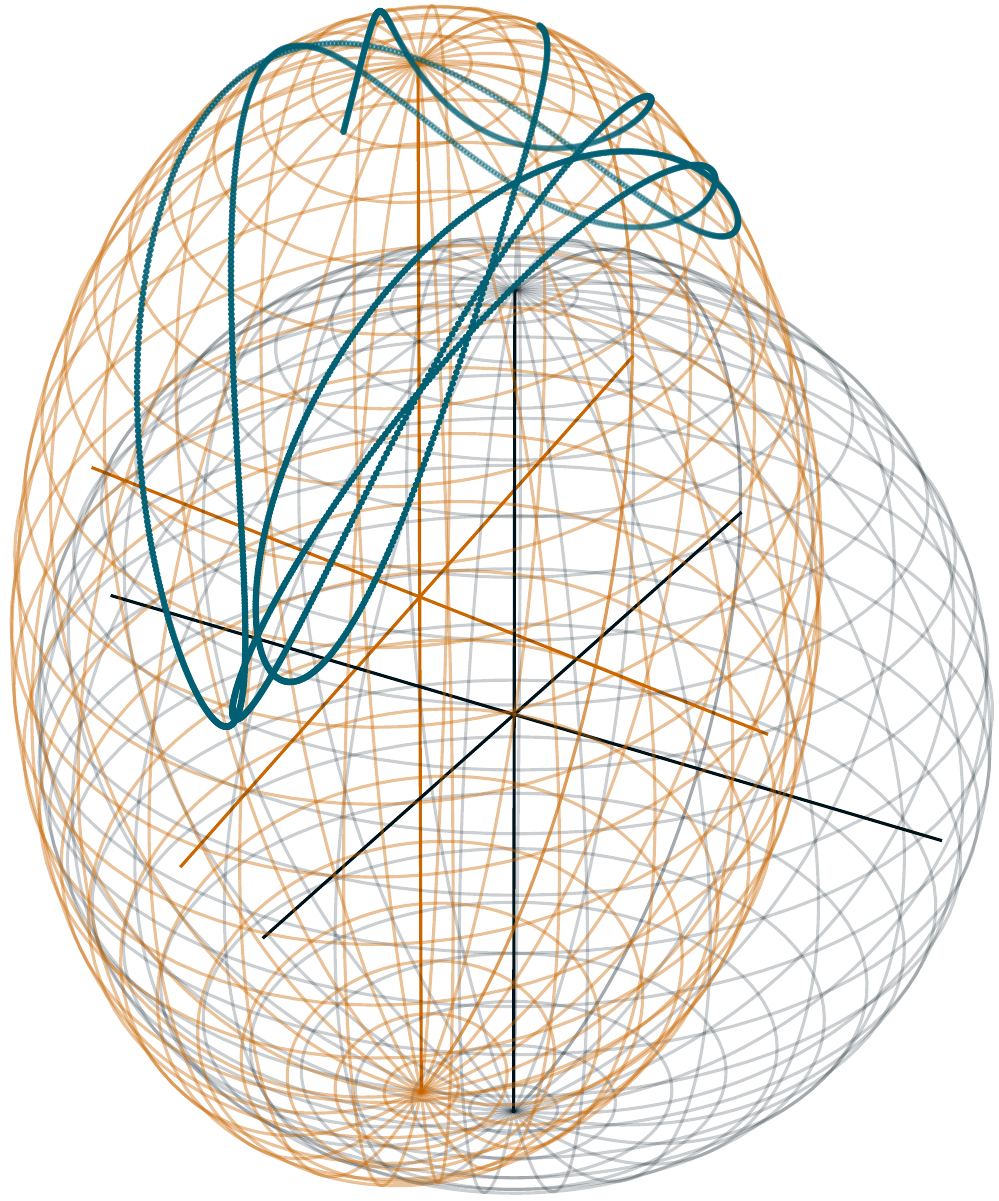}%
\label{fig: sim_data_mid_plot}}
\hfil
\subfloat[]{\includegraphics[width=1.05in]{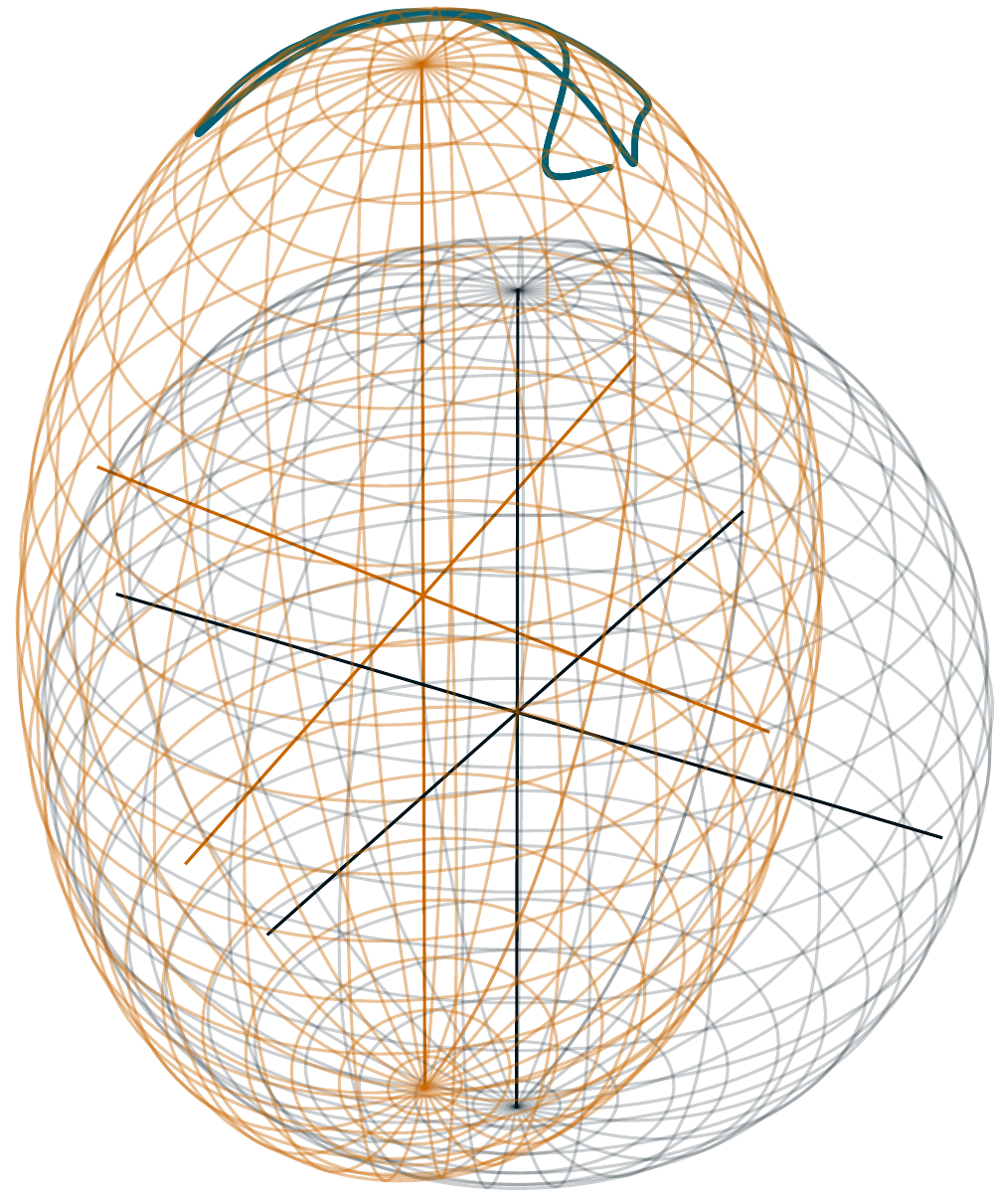}%
\label{fig: sim_data_low_plot}}
\caption{Simulated magnetometer data for three datasets: (a) WAM, (b) MAM, and (c) LAM. The 3D plots show blue dots for magnetometer data, gray spheres for the true magnetic field, and orange ellipsoids for the distorted magnetic field.}
\label{fig_sim_data_plots}
\end{figure}
\section{Evaluation Methodology}
\label{sec:evaluation}

\noindent We compared the performance of five methods for three-axis magnetometer calibration and, optionally, for three-axis gyroscope calibration. These methods can be divided into batch and real-time solutions, with the proposed methods highlighted in bold. The batch methods are as follows:

%
\begin{enumerate}
\item \textit{\textbf{\MAGYCBFG}}: The calibration parameters are estimated using the batch mode factor graph approach described in Section \RomanNumeralCaps{3}.C, where all the factors are added to the factor graph before optimization.
\item \textit{\TWOSTEP}: The calibration parameters are estimated using the widely cited \TWOSTEP method \cite{Alonso2002b}, which is based on the implementation proposed by Dinale \cite{Dinale2013}. This method takes as input the local value of the magnetic field.
\item \textit{Ellipsoid Fit}: The calibration parameters are estimated using the widely used Ellipsoid Fit method, based on the implementation proposed by Bazhin et al. \cite{Bazhin2022}.
\end{enumerate}

\noindent The real-time methods are listed below:

%
\begin{enumerate}
\item \textit{\textbf{\MAGYCIFG}}: The calibration parameters are estimated using the incremental mode factor graph approach described in Section \RomanNumeralCaps{3}.C, where the factors are added as they are received.
\item \textit{\MagFactor}: The calibration parameters are estimated using a factor graph approach that utilizes the magnetic calibration factor provided in the GTSAM library \cite{BORGLab2023}. Unlike full soft-iron matrix estimation, this method only estimates a single scale factor that is uniform across all three axes, as well as the hard-iron bias. The current attitude of the system and the local magnetic field value must be provided as inputs. To the best of our knowledge, this is the only factor graph-based method available in the literature that accounts for soft- and hard-iron biases and does not require a predefined set of movements.
\end{enumerate}

To compare batch and real-time methods, the calibration parameters estimated with the real-time methods were based on the average of the last \SI{20}{\percent} of the estimated parameters. It should be noted that the \TWOSTEP and \MagFactor methods require knowledge of the local magnetic field magnitude, which was obtained from the World Magnetic Model provided by the \ac{noaa} \cite{NOAA2023} for in-field evaluations. Furthermore, for these methods, we assume the gyroscope is calibrated; however, this has no impact on the methods' performance or evaluation, as neither of them uses the angular rate measurements nor computes the gyroscope bias.

For both numerical and in-field evaluations, for the factor graph-based methods, the termination criteria were also set empirically, with both the relative and absolute error tolerances set to $1.0 \times 10^{-7}$. A multifrontal Cholesky factorization was used, as it has been shown to outperform the LDL and QR factorizations \cite{Dellaert2006}. The optimization was computed using the RISE method from GTSAM. The initial conditions for the state vector are the assumption of both magnetometer and gyroscope to be calibrated, i.e., \mbox{$\mathbf{m_b} = \SI[parse-numbers=false]{\vec{\mathbf{0}}}{\milligauss}$}, \mbox{$\mathbf{w_b} = \SI[parse-numbers=false]{\vec{\mathbf{0}}}{\milli\radian\per\second}$} and \mbox{$\mathbf{a} = ( 1.0 \; 0.0 \; 0.0 \; 1.0 \; 0.0 \; 1.0 )^T$}.

For \MAGYCIFG and \MAGYCBFG, the factor graph was optimized each time a new pair of factors was added from an averaged sample set $i, \;\forall \; i \; \in \; \{0, \dots, n\}$, with an average window size $\theta$ set to match the sensor's frequency. The noise covariance matrices were set to $\mathbb{I}_3 \cdot 0.001$ and $\mathbb{I}_1 \cdot 0.01$ for (\ref{eq: unary_factor_residual}) and (\ref{eq: unary_factor_norm}), respectively. These values were determined through a sensitivity analysis, which found the optimal trade-off for the matrices. Due to space constraints, the details of this analysis are not included.
\section{Numerical Simulation Evaluation}
\label{sec:sim_results}

%
\noindent A Monte Carlo numerical simulation was conducted in \textit{Python3} running on a 12th-generation Intel Core i7-12800H CPU with 64 GB of memory to replicate 10,000 measurements from a \ac{mems} \ac{ahrs} during sinusoidal motions of a vehicle. Three simulated datasets represented varying degrees of angular motion constraint in all degrees of freedom. The wide angular movement (WAM) dataset (Fig. \ref{fig: sim_data_high_plot}) covered a $\pm \SI{180}{\degree}$ range in roll, pitch, and heading. The moderate angular movement (MAM) dataset (Fig. \ref{fig: sim_data_mid_plot}) had $\pm \SI{5}{\degree}$ and $\pm \SI{45}{\degree}$ ranges for roll and pitch, respectively, but the same heading range as WAM. The low angular movement (LAM) dataset (Fig. \ref{fig: sim_data_low_plot}) retained the roll range while reducing the pitch and heading ranges to $\pm \SI{15}{\degree}$ and $\pm \SI{90}{\degree}$, respectively. Each experiment lasted \SI{400}{\second}, with simulated data generated at a \SI{25}{\hertz} rate and magnetometer measurements ($\sigma_{mag} = \SI{1}{\milligauss}$) and angular rate sensor ($\sigma_{gyro} = \SI{5}{\milli\radian\per\second}$) corrupted by Gaussian noise.

The true magnetic field vector is given by $\SI[parse-numbers=false]{\mathbf{m_0} = [227,\, 52, \,412]^T}{\milligauss}$, the soft-iron upper triangular terms are given by $\mathbf{a} = [1.10,\, 0.10,\, 0.04,\, 0.88,\, 0.02,\, 1.22]^T$, the hard-iron bias is $\SI[parse-numbers=false]{\mathbf{m_b} = [20,\, 120,\, 90]^T}{\milligauss}$, and the gyroscope bias is $\SI[parse-numbers=false]{\mathbf{w_b} = [4,\, -5,\, 2]^T}{\milli\radian\per\second}$. To account for the magnetic field's uncertainty from local perturbation, in the TWOSTEP and MagFactor3 methods it was \SI{5}{\percent} higher than the value used to generate the simulated data.

%
\begin{table*}[!t]
\renewcommand{\arraystretch}{1.4}
\caption{Mean heading RMSE and magnetic field standard deviation for three batch and two real-time calibration methods and three simulated datasets over 100 validation simulations. N/A indicates failure. The best two results in each column are bolded}\label{table: simulation_results}
\centering
\begin{tabular}{p{0.05cm} p{0.1cm} l c@{\hspace{5mm}}c c@{\hspace{5mm}}c c@{\hspace{5mm}}c c@{\hspace{5mm}}c}
\multicolumn{2}{c}{\multirow{4}[4]{*}{}} &
\multicolumn{1}{c}{\multirow{4}[4]{*}{}} &
\multicolumn{2}{c}{\textbf{WAM for Calibration}} &
\multicolumn{2}{c}{\textbf{MAM for Calibration}} &
\multicolumn{2}{c}{\textbf{LAM for Calibration}}  \\
\Xhline{1.5pt}
& & & Mean Heading & Magnetic Field & Mean Heading & Magnetic Field & Mean Heading & Magnetic Field \\
& & & RMSE (deg) & Std (mG) & RMSE (deg) & Std (mG) & RMSE (deg) & Std (mG) \\
\hline
\multicolumn{2}{c}{}  & Raw & 28.864 & 60.330 & 28.864 & 60.330 & 28.864 & 60.330 \\
\hline
\multicolumn{2}{c}{\parbox[t]{2mm}{\multirow{3}{*}{\rotatebox[origin=c]{90}{BATCH}}}} & \MAGYCBFG & 2.659 & 9.794 & \textbf{2.735} & \textbf{9.875} & \textbf{3.812} & \textbf{16.016} \\

\multicolumn{2}{c}{}                                      & \TWOSTEP & \textbf{2.518} & \textbf{9.421} & N/A & N/A & N/A & N/A \\

\multicolumn{2}{c}{}                                      & Ellipsoid Fit & 18.232 & 62.522 & 16.235 & 60.211 & 50.842 & 157.802 \\
\hline
\parbox[t]{2mm}{\multirow{2}{*}{\rotatebox[origin=c]{90}{REAL}}} & \parbox[t]{2mm}{\multirow{2}{*}{\rotatebox[origin=c]{90}{TIME}}} & \MAGYCIFG & \textbf{2.654} & \textbf{9.762} & \textbf{2.791} & \textbf{10.371} & \textbf{3.528} & \textbf{17.552} \\
\multicolumn{2}{c}{}                                             & \MagFactor & 15.915 & 45.192 & 16.917 & 49.603 & 14.431 & 48.784 \\
\Xhline{1.5pt}
\end{tabular}
\end{table*}

%
The calibration methods presented in Section \RomanNumeralCaps{4} were calibrated on the three datasets mentioned previously and later evaluated on a dedicated unique excited evaluation dataset to avoid overfitting. Results in Table \ref{table: simulation_results} and Fig. \ref{fig: sim_matrix_error} show that the proposed \MAGYC methods consistently outperformed the benchmark methods. Specifically, in the WAM dataset calibration, while the \TWOSTEP method showed superior performance in hard-iron estimation, both \MAGYCBFG and \MAGYCIFG displayed comparable performance in mean heading \ac{rmse} and magnetic field standard deviation.

In the MAM and LAM datasets, \TWOSTEP failed to converge, and the Ellipsoid Fit method significantly deteriorated results, highlighting their limitation to wide-range movement scenarios, which may not always be feasible in some full-scale vehicles. Meanwhile, \MagFactor showed more robustness but did not compute non-orthogonality or scale factors in the three axes. In contrast, the proposed \MAGYC methods consistently demonstrated better overall performance.

These results suggest that the \MAGYC methods are more resilient to constrained angular movements than state-of-the-art methods, particularly in the MAM and LAM datasets that emulate actual operating conditions. Furthermore, the \MAGYC methods show consistent performance even under conditions favorable to the benchmark methods, such as in the WAM dataset. Additionally, the \MAGYCBFG method demonstrates greater robustness compared to \MAGYCIFG. Furthermore, in the case of the real-time \MAGYCIFG method, convergence to a constant bias estimation occurs after analyzing \SI{40}{\percent} of the data.

Regarding processing time, the \MAGYCIFG method has a computation time of around \SI{0.8}{\second} per calibration for all datasets, demonstrating the feasibility of real-time operation when the averaging window ($\theta$) is set to the sensor's frequency, as proposed in Section \RomanNumeralCaps{3}.C, with a factor graph update rate of \SI{1}{Hz} (one update per second). In contrast, the \MAGYCBFG method shows a slight increase in processing time with a decrease in the movement range, increasing from around \SI{0.3}{\second} to \SI{0.8}{\second} per calibration due to longer iterations required to meet termination criteria; however, this is not critical for operation due to the post-processing use of the \MAGYCBFG method.

These simulation results support that both proposed \MAGYC methods show competitive or better performance than the benchmark methods, indicating their effectiveness for post-processing and real-time navigation applications. Furthermore, even though the proposed methods do not significantly improve performance for the WAM dataset, they demonstrate better performance in movement-constrained scenarios (MAM and LAM).
\section{Field Experimental Evaluation}
\label{sec:field_results}


\begin{figure*}[!t]
\centering
\subfloat[]{\includegraphics[width=4.5in]{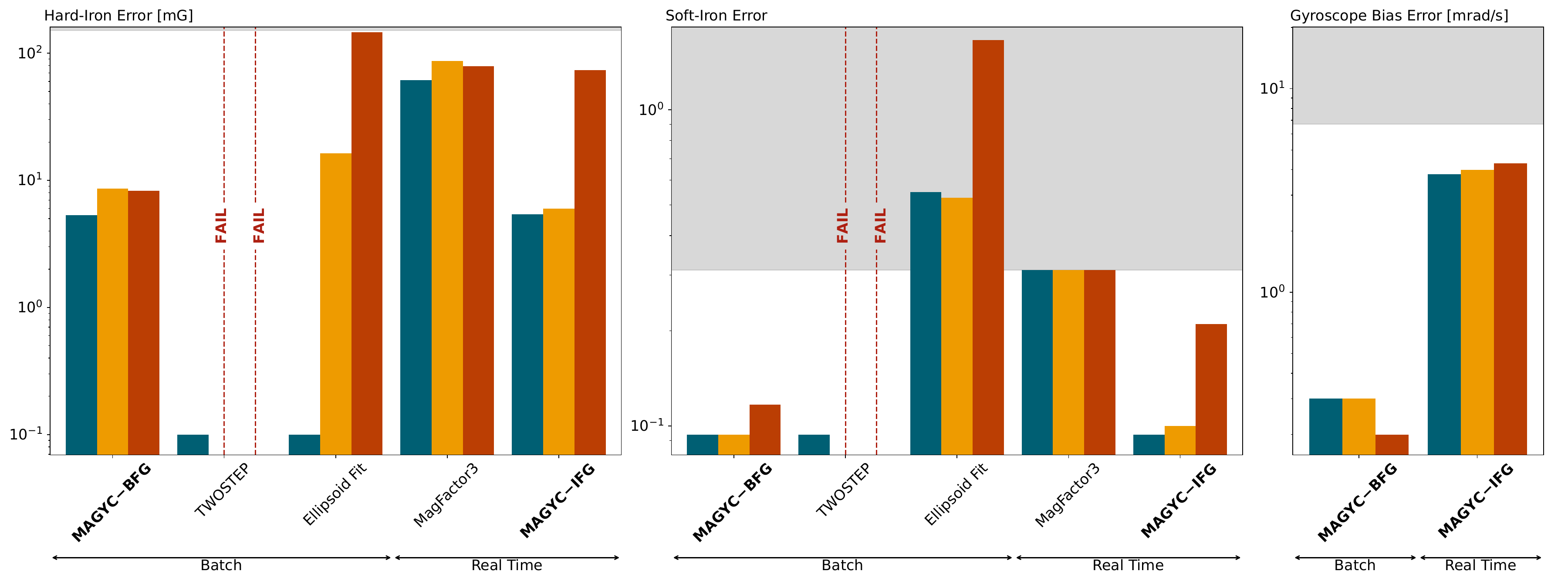}%
\label{fig: sim_matrix_error}}
\hfil\hfil\hfil
\subfloat[]{\includegraphics[width=2.3in]{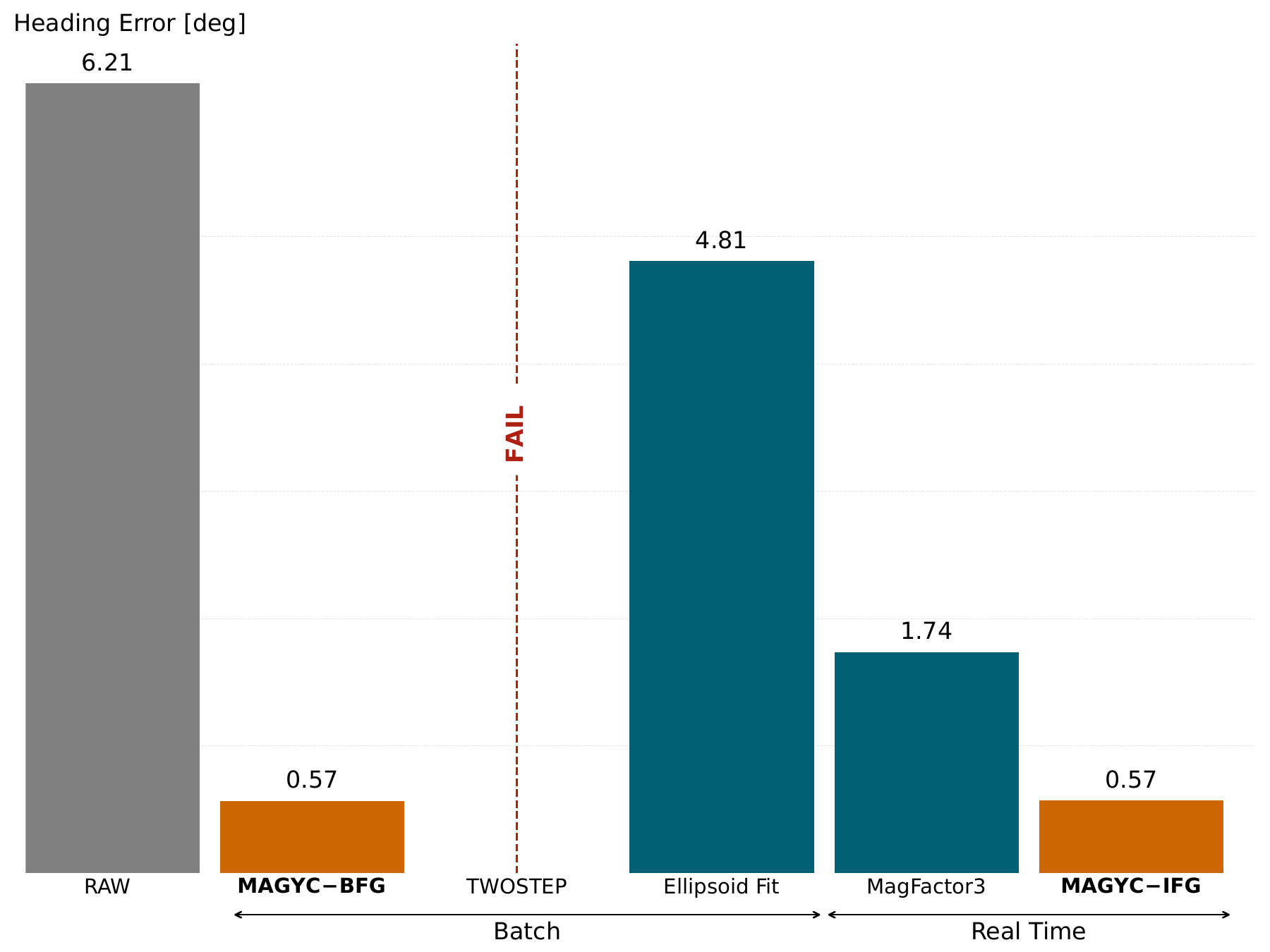}%
\label{fig: field_heading_error}}
\caption{Evaluation results: (a) Performance comparison of five calibration methods on three simulated datasets. The hard-iron error $|m_b - m_b^*|$, soft-iron error $|A - A^*|$, and gyroscope bias error $|w_b - w_b^*|$ are analyzed for the WAM (green), MAM (yellow), and LAM (orange) datasets. Red dashed lines indicate instances where the method failed to estimate the parameters for a particular dataset, and gray-shaded zones, show the raw data value. (b) Heading error on field data for calibration parameters estimated with EXP1 and evaluated with EXP2.}
\label{fig: field_results}
\end{figure*}

\begin{figure}[!b]
\centering
\subfloat[]{\includegraphics[width=1.4in]{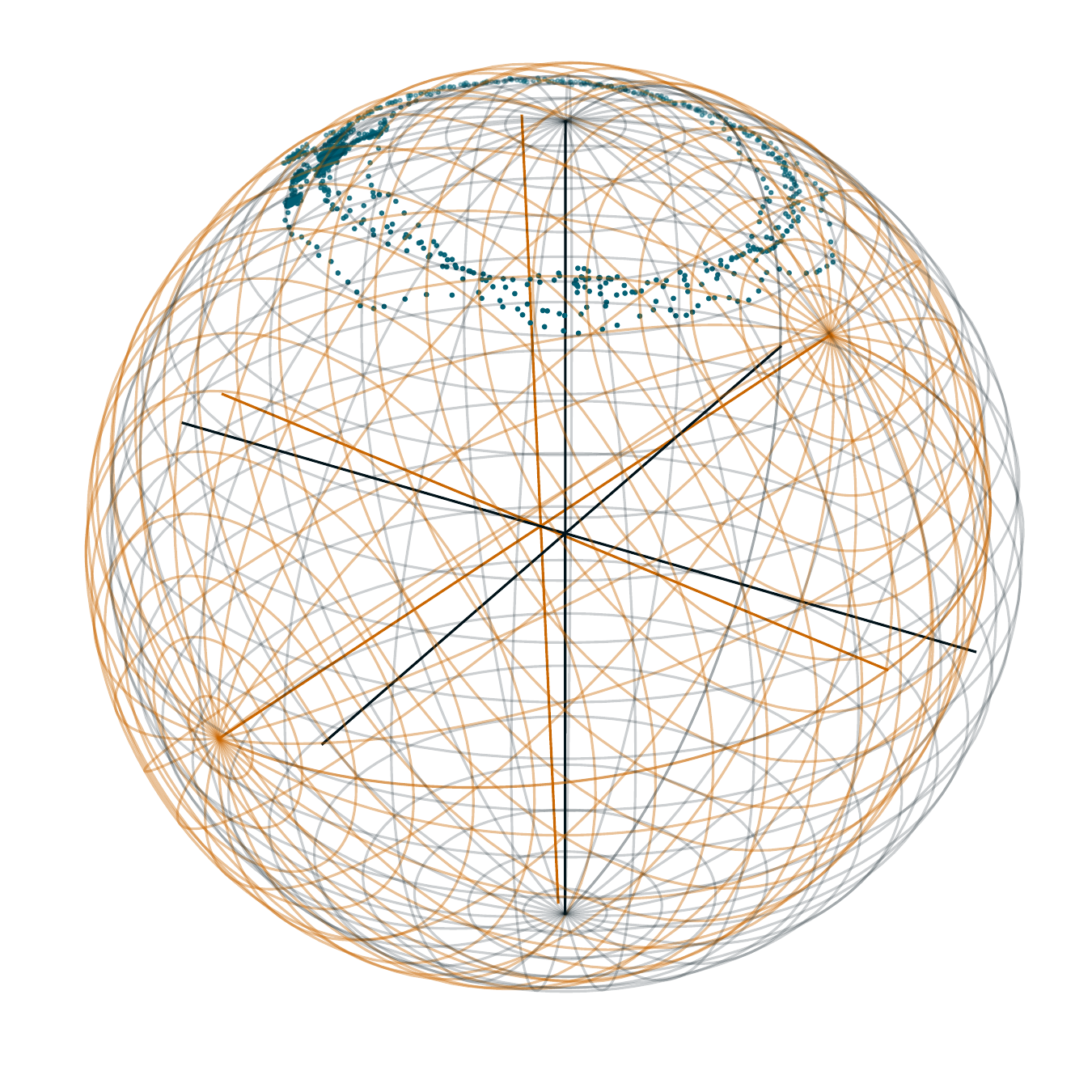}%
\label{fig: exp1_plot}}
\hfil
\subfloat[]{\includegraphics[width=1.4in]{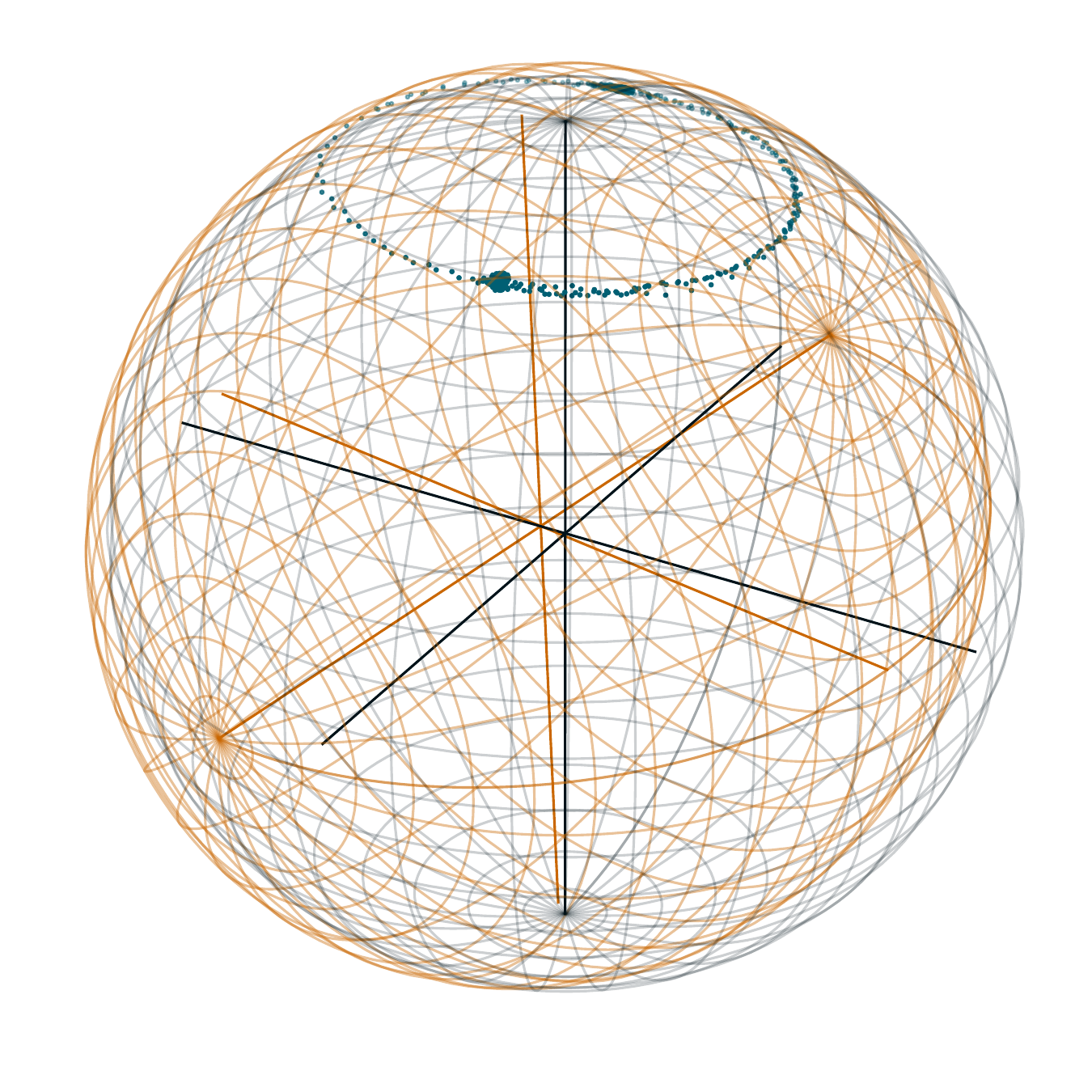}%
\label{fig: exp2_plot}}
\caption{In-field magnetic data for two datasets: (a) EXP1 and (b) EXP2. The 3D plots show blue dots for magnetometer data and gray spheres for true magnetic field.}
\label{fig_exp_data_plots}
\end{figure}

%
\noindent We evaluate the in-field performance of the proposed and benchmark methods using a seafloor mapping survey dive conducted during an oceanographic survey mission in Monterey Bay, where the local magnetic field had a magnitude of \SI{479}{\milligauss} \cite{NOAA2023}, by the Doc Ricketts \ac{rov}, owned and operated by the Monterey Bay Aquarium Research Institute (MBARI), which is equipped with a \ac{mems} \ac{imu}. Two field experiments were conducted on the same day. The first experiment denoted as EXP1 (Fig. \ref{fig: exp1_plot}), is a magnetometer calibration procedure for \acp{rov}, which involved a series of \SI{360}{\degree} heading rotations, with the pitch and roll configurations changing to produce \SI{5}{\degree} pitch and roll movements. The second experiment denoted as EXP2 (Fig. \ref{fig: exp2_plot}), consisted of a standard survey where the vehicle maintained a stable pitch and roll, resembling the pattern of ``mowing a lawn''.

%
\noindent To evaluate the heading estimation performance, we used a Vectornav VN100 MEMS-based IMU operating at a sampling rate of \SI{80}{Hz}, featuring a magnetometer noise level of $\sigma_{mag} = \SI{1}{\milligauss}$ and an angular-rate gyroscope noise level of $\sigma_{gyro} = \SI{0.5}{\milli\radian\per\second}$ \cite{vn100}. In field applications, although magnetometers can be calibrated beforehand, once the sensor is mounted on the vehicle, local perturbations can cause unknown magnetometer biases. Therefore, we utilized a Kearfott SeaDeViL high-end \ac{ins} operating at a sampling rate of \SI{25}{\hertz} as the ground truth for heading comparison. The \ac{ins} includes a ring-laser gyro, providing a precision of \SI{0.05}{\degree} and \SI{0.03}{\degree} in heading and pitch/roll, respectively \cite{kearfott}.  We interpolated the \ac{mems} \ac{imu} data to the \ac{ins} sampling time to estimate the vehicle's heading. The heading error, defined as the standard deviation between the measured heading from the INS and the calculated heading from the bias-compensated magnetometer data for each evaluated method, was used as the evaluation metric in order to isolate alignment errors.

As shown in Fig. \ref{fig: field_heading_error}, the proposed \MAGYC methods demonstrate remarkable performance when calibrated using EXP1 and evaluated with EXP2, reducing the original heading error from \SI{6.21}{\degree} to \SI{0.57}{\degree}. Furthermore, the proposed methods improve the heading error for calibration with EXP2, while the benchmark methods fail to converge to a solution except for \MagFactor, which is still outperformed by both proposed methods, thereby demonstrating the superiority of the proposed methods. The performance of the benchmark methods highlights how these methods are constrained to wide-range movement applications, rendering them unsuitable for full-scale applications. This underscores the capability of the proposed methods for enhancing calibration in highly constrained implementations, as they consistently converge even in challenging scenarios, whether in batch or real-time mode.

\addtolength{\textheight}{-1.0cm}

\section{Conclusions}
\label{sec:conclusion}

%

\noindent The MAgnetometer and GYroscope Calibration (\MAGYC) proposed methods, \MAGYCBFG and \MAGYCIFG, have been demonstrated to significantly improve the performance of AHRS sensors in simulated and in-field scenarios. The results show that the \MAGYC methods outperform or are comparable to previously reported methods, such as \TWOSTEP, Ellipsoid Fit, and \MagFactor. Notably, in scenarios where the sensor ranges are constrained, such as the simulated datasets MAM and LAM or the field data, \TWOSTEP failed to converge while Ellipsoid Fit provided inaccurate calibration, even when precise knowledge of the local magnetic field vector was available in both cases, rendering them unsuitable for full-scale vehicles where extensive angular motions in the three rotational degrees of freedom are unfeasible. Although in these cases \MagFactor converged, it did not compute non-orthogonality or scale factors in the three axes and was consistently outperformed even when accurate knowledge of the local magnetic field and system attitude was required.

In all scenarios, \MAGYC methods converged and improved the calibration of the sensors. This underscores the capability of the proposed methods for enhancing calibration in highly constrained implementations, as they consistently converge even in challenging scenarios, whether in batch or real-time mode.

These results highlight the practicality and efficacy of the \MAGYC methods proposed in this study for the calibration of magnetometers and gyroscopes, particularly in the context of attitude estimation. The implications of these findings extend to potential advancements in cost-effective navigation systems and enhanced performance for ground, marine, and aerial vehicles in real-world environments.

\section*{Acknowledgments}

The field experimental data used in this study were collected during the 2014 Ocean Imaging cruise conducted by MBARI, led by Chief Scientist Dr. David Caress. Additionally, we acknowledge Dinale \cite{Dinale2013} and Bazhin et al. \cite{Bazhin2022} for providing open-source code for the implementation of the \TWOSTEP \cite{Alonso2002b} and Ellipsoid Fit methods, respectively.

\bibliography{refs/IEEEabrv,refs/MagCal}
\bibliographystyle{IEEEtran}

\end{document}